\documentclass[conference]{IEEEtran}

\usepackage{cite}
\usepackage{graphicx}
\usepackage{amsmath}
\usepackage{algorithmic}
\usepackage{algorithm}
\usepackage{subfig}
\usepackage[percent]{overpic}
\usepackage{color}

\begin{document}
% paper title
\title{Parameter Identification of Induction Motor Using Modified Particle Swarm Optimization Algorithm}

% Author names and affiliations
\author{%
\authorblockN{Hassan M. Emara}
\authorblockA{Department of Electrical\\
              Power and Machines,\\
              Faculty of Engineering,\\
	      Cairo University\\
	      \texttt{hmrashad@ieee.org}
              }
\and
\authorblockN{Wesam Elshamy}
\authorblockA{Department of Computing\\
              and Information Sciences,\\
              Kansas State University\\
              \texttt{welshamy@ksu.edu}}
\and
\authorblockN{A. Bahgat}
\authorblockA{Department of Electrical\\
              Power and Machines,\\
              Faculty of Engineering,\\
	      Cairo University\\
              \texttt{ahmed.bahgat@ieee.org}}}

\maketitle

\begin{abstract}
 This paper presents a new technique for induction  motor parameter identification. The proposed technique is based on a simple startup test using a standard V/F inverter. The recorded startup currents are compared to that obtained by simulation of an induction motor model.  A Modified PSO optimization is used to find out the best model parameter that minimizes the sum square error between the measured and the simulated currents. The performance of the modified PSO is compared with other optimization methods including line search, conventional PSO and Genetic Algorithms. Simulation results demonstrate the ability of the proposed technique to capture the true values of the machine parameters and  the superiority of the results obtained using the modified PSO over other optimization techniques.
\end{abstract}

\section{Introduction}
Induction motors are the most widely used motors in industry because they are simple to build, rugged, reliable and have good self-starting capability. The majority of control schemes of such motor drives require exact knowledge of at least some of the induction motor parameters. Mismatch between the actual motor parameter values and that used within the controller leads to deterioration in the drive performance \cite{vas98}. In order to avoid performance degradation, motor drives usually perform a pre-tuning algorithm during inverter initialization. The pre-tuning is based on offline parameter estimation using data available from simple test of motor performance while supplied by the inverter. Several methods have been proposed to tackle the problem of offline induction machine parameter estimation \cite{tol02}.

The rapidly increasing computational power of personal computers allowed researchers to implement several optimization algorithms and verify their efficiency. Researchers developed many algorithms that mimic natural phenomena. Examples of these algorithms include the Simulated Annealing \cite{kirk83}, Genetics Algorithms (GA) \cite{hol75},  Ant Colony Optimization \cite{col91} algorithms.

Particle Swarm Optimization (PSO) \cite{ken95} is among these nature inspired algorithms. It is inspired by the ability of birds flocking to find food that they have no previous knowledge of its location. Every member of the swarm is affected by its own experience and its neighbors' experiences. Although the idea behind PSO is simple and can be implemented by two lines of programming code, the emergent behavior is complex and hard to completely understand \cite{ken04}. 

In this paper different versions of PSO are used to identify six parameters of the motor. The results obtained using these optimizers are presented and discussed.

\section{PARTICLE SWARM OPTIMIZATION}
Particle Swarm Optimization was inspired by the ability of a flock of birds or a school of fish to capitalize on their collective knowledge in finding food or avoiding predators. Each swarm member or particle has a small memory that enables it to remember the best position it found so far and its goodness. Particles are affected by their own experience (best found position) and their neighbors' experiences (best found position by the neighbors). The behavior of the particles is described by \eqref{pso_v} and \eqref{pso_x}.
\begin{multline}
  \label{pso_v}
  v_{id}(t+1) = w \times v_{id}(t) + lrn_1 \times rand_1 \times (p_{id}(t) - x_{id}(t))\\
  + lrn_2 \times rand_2 \times (p_{gd}(t) - x_{id}(t))    
\end{multline}
\begin{equation}
  \label{pso_x}
  x_{id}(t+1) = x_{id}(t) + v_{id} (t+1)
\end{equation}
In \eqref{pso_v}, $v_{id}$ is the speed of particle $i$ in dimension $d$. The first right hand side term corresponds to the inertia force that pushes the particle in its old direction, where $w$ is the weight value that controls this inertia force. The second term corresponds to the cognitive or personal experience component. It attracts the particle from its current position $x_{id}$ to its best found position so far in that dimension $p_{id}$ affected by a learning weight $lrn_1$ and a uniformly distributed random variable $rand_1$ in the range $(0, 1)$.  The third term corresponds to the social influence of the neighbors on the particle. It affects the particle by attracting it from its current position $x_{id}$ to the best position found by its neighbors $p_{gd}$ and this influence is controlled by a learning weight $lrn_2$ and another independent random variable $rand_2$ uniformly distributed in the range $(0, 1)$. For each time step, as described by \eqref{pso_x}, each particle moves by a step of value $v_{id}$ in the $d^{th}$ dimension.

The PSO algorithm itself has evolved. The weight parameter $w$ was not included in the basic algorithm. It was added later and researchers examined the effect of varying its value \cite{shi98}. A speed limit for the particles was introduced to prevent the explosion of speed values.

PSO operates in three spaces, the social network space, the parameter space of problem variables and the evaluative space \cite{ken04} where estimates for the goodness of solutions are defined. Various social networks have been proposed and investigated by researchers \cite{ken99}. In the original PSO algorithm, the social network connects every particle to all other particles and it is only influenced by the one that has the best experience compared to all particles. We will refer to this algorithm as PSO-g where `g' stands for \emph{global}. Though this algorithm converges rapidly, it could get easily trapped in local minima. 

A variant of the simple PSO has a ring social network. In this algorithm the particles are arranged in an imaginary ring and every particle is connected to its immediately preceding and succeeding particles in this ring. We will refer to this algorithm as PSO-l where `l' stands for \emph{local}. This algorithm converges slower than PSO-g but it is less susceptible to local minima and enjoys a higher degree of particles diversity. The influence of each particle in the swarm is limited to its two immediate neighbors. This influence limitation helps the particles to explore the search space with different points of attraction instead of a single best found point in the PSO-g algorithm. On the other hand, it may lead to excessive wandering for the particles leading to slow convergence even in easy problems having single optimum.

Both PSO-g and PSO-l are based on a static neighborhood network. The first stages of the search for the global best position require exploration of possible solutions, which PSO-l can do better. Later stages require exploitation of the best found candidate solutions by early stages of the search, which PSO-g is clever at. Hence, researchers suggested using a dynamic neighborhood.

In \cite{sug99}, the neighborhood of each swarm member expands from an initial network that connects each particle to itself at early stages of the search, to a network that fully connects it to all other particles. This algorithm transforms gradually from acting like PSO-l in early stages of the search, to behave more like PSO-g at late stages. Two network expanding procedures have been introduced. Both of them depend on the current position of the particle to search for nearby particles to add to its neighborhood list.

In \cite{veer03}, a Fitness-Distance-Ratio based PSO (FDR-PSO)  algorithm is introduced. In this algorithm, each particle is affected by three components; the cognitive, social and the FDR components. The third component corresponds to the influence of the particle that maximizes the FDR. The higher the fitness of the neighbor and the closer its distance to the original particle, the more likely it will influence this particle. A new learning factor is introduced for the FDR component.

In \cite{moh05}, a randomly generated directed graphs are used to define neighborhood where graph links are unidirectional. Two methods for modifying the neighborhood structure are tested. The `random edge migration' method disconnects one side of an edge and connects it to another neighbor, while the `neighborhood re-structuring' method totally re-initializes the structure after it is kept fixed for a period of time.

In \cite{jan05}, a Hierarchical PSO (H-PSO) version is introduced. In this algorithm, particles are arranged in a hierarchy structure and the best performing particles ascend the tree to influence more particles, replacing relatively worse performing particles which descend the tree. A variant of this algorithm where the structure of the tree itself is made dynamic is presented and tested.

\section{CLUBS-BASED PSO}
PSO first models were confined to perceive the swarm as a flock of birds that fly in the search space. The picture of flying birds has limited the imagination of researchers somehow for sometime. Recently, a more broad perception of the swarm as a group of particles, whether birds, humans, or any socializing group of particles began to emerge.

In our proposed Clubs-based PSO (C-PSO) algorithm, we create \emph{clubs} for particles analogous to our clubs where we meet and socialize. In our model, every particle can join more than one club, and each club can accommodate any number of particles. Vacant clubs are allowed.

After randomly initializing the particles position and speed in the initialization range, each particle joins a predefined number of clubs, which is known as its \emph{default membership level}, and the choice of these clubs is made random. Then, current values of particles are evaluated and the best local position for each particle is updated accordingly. While updating the particles' speeds, each particle is influenced by its best found position and the best found position by all its neighbors, where its neighborhood is the set of all clubs it is a member of. After speed and position update, the particles' new positions are evaluated and the cycle is repeated.

While searching for the global optimum, if a particle shows superior performance compared to other particles in its neighborhood, the spread of the strong influence by this particle is reduced by reducing its membership level and forcing it to leave one club at random to avoid premature convergence of the swarm. On the other hand, if a particle shows poor performance, that it was the worst performing particle in its neighborhood, it joins one more club selected at random to widen its social network and increase the chance of learning from better particles. 

The cycle of joining and leaving clubs is repeated every time step, so if a particle continues to show the worst performance in its neighborhood, it will join more clubs one after the other until it reaches the maximum allowed membership level. While the one that continues to show superior performance in every club it is a member of will shrink its membership level and leave clubs one by one till it reaches the minimum allowed membership level.

During this cycle of joining and leaving clubs, particles which no longer show extreme performance in their neighborhood, either by being the best or the worst, go back gradually to default membership level. The speed of going back to default membership level is made slower than that of diverting from it due to extreme performance. The slower speed of regaining default membership level allows the particle to linger, and adds some stability and smoothness to the performance of the algorithm. A check is made every $rr$ (retention ration) iterations to find the particles that have membership levels above or below the default level, and take them back one step towards the default membership level if they do not show extreme performance.

The proposed algorithm can be described by Algorithm \ref{cpso_pseudo}

\begin{algorithm}
\caption{C-PSO pseudocode \label{cpso_pseudo}}
\begin{algorithmic}
  \STATE \textbf{begin}
  \STATE Initialize particles and clubs
  \WHILE{(termination condition = \texttt{false})}
    \STATE evaluate particles fitness: $f(x)$
    \STATE update $P$
    \FOR{$i=1$ to number of particles}
      \STATE $g_i$ = best of $neighbors_i$
      \FOR{$d=1$ to number of dimensions}
        \STATE $v_{id} = w \times rand_1 \times v_{id} + lrn_1 \times rand_2 \times (p_{id} - x_{id}) + lrn_{2} \times rand_3 \times (g_{id} - x_{id})$
        \STATE $x_{id} = x_{id} + v_{id}$
      \ENDFOR
    \ENDFOR
    \STATE update neighbors
    \FOR{$j=1$ to number of particles}
      \IF{($x_j$ is best of $neighbors_j$) \textbf{and} ($|membership_j|$ $>$ $min\_membership$)}
        \STATE leave random club
      \ENDIF
      \IF{($x_j$ is worst of $neighbors_j$) \textbf{and} ($|membership_j|$ $<$ $max\_membership$)}
        \STATE join random club
      \ENDIF
      \IF{($|membership_j| \neq default\_membership$) \textbf{and} (remainder($iteration/rr$) = 0)}
        \STATE update $membershhip_j$
      \ENDIF
    \ENDFOR
    \STATE $itermation = iteration + 1$
    \STATE evaluate termination condition
  \ENDWHILE
\end{algorithmic}
\end{algorithm}

\begin{figure}
  \centering
  \includegraphics[width = 0.7\linewidth]{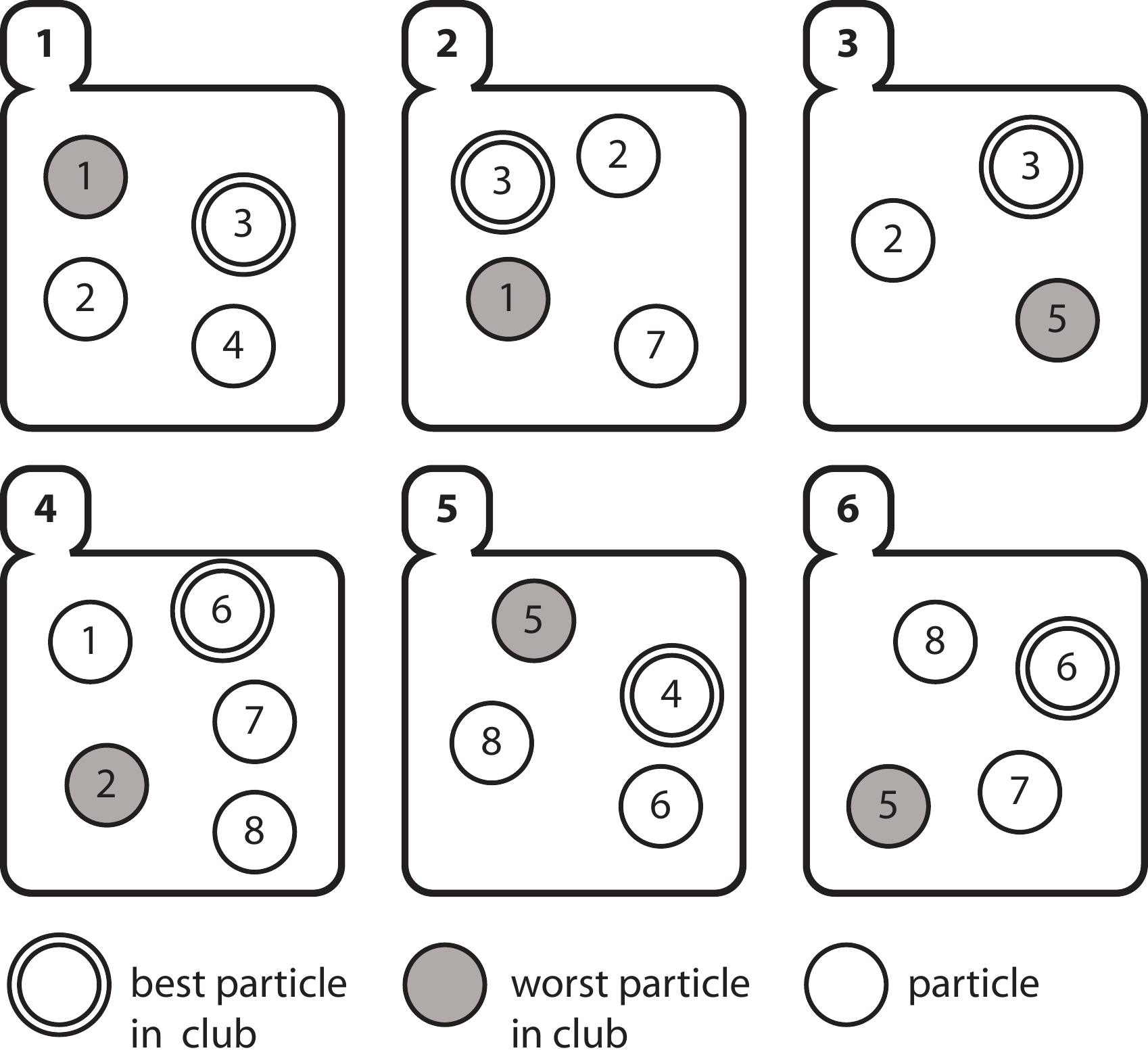}
  \caption{A snapshot of clubs during a simulation of the C-PSO algorithm \label{pso-snap}}
\end{figure}

Where $P$ is local best position, $neighbors_i$ is the set of particle $i$ neighbors, $membership_i$, $|membership_i|$ are the set of clubs that particle $i$ is a member of and the size of this set, respectively. $rand_{1,2,3}$ are three independent uniformly distributed random numbers in the range (0,1).  Fig. \ref{pso-snap}  shows a snapshot of the clubs during an execution of the C-PSO algorithm.  In this example, the swarm consists of 8 particles, and there are 6 clubs available for them to join.

Given the previous pseudocode, and that the minimum, default and maximum membership levels are 2, 3 and 5 respectively, the following changes in membership will happen to particles in Fig. \ref{pso-snap} for the next iteration which is a multiple of $rr$:
\begin{enumerate}
 \item particle 3 will leave club 1, 2 or 3 because it is the best particle in its neighborhood.
 \item Particle 5 will join club 1, 2 or 4 because it is the worst particle in its neighborhood.
 \item Particle 2 will leave club 1, 2, 3 or 4, while particle 4 will join club 2, 3, 4, or 6 to go one step towards default membership level because they do not show extreme performance in their neighborhood.
\end{enumerate}

\section{INDUCTION MOTOR PARAMETER IDENTIFICATION}
\subsection{Induction Motor Model}
\begin{figure}
  \centering
  \resizebox{0.9\linewidth}{!}{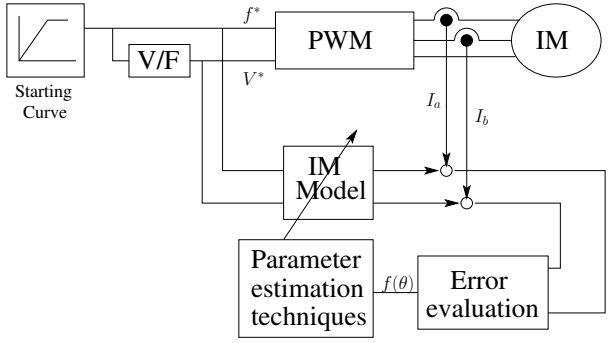}
  \caption{Parameter identification using $V/F$ starting of induction machine \label{ind_model}}
\end{figure}

In this paper, parameter identification of the induction machine involves the estimation of the induction machine, namely: stator resistance, rotor resistance, Leakage inductances of the stator and rotor, mutual inductance, and equivalent rotor inertia.  As indicated in Fig. \ref{ind_model}, the proposed test is based on a simple startup via either direct on line or constant $V/F$ converter. The varying frequency excites the different motor dynamics, while the constant $V/F$ keeps the machine flux nearly constant and equal to the rated flux.  The starting current wave is recorded and several identification algorithms are used to find out appropriate parameter values that can minimize the integrated absolute error between the recorded waveform and that generated by a motor model using the identified parameters.

For the purpose of this study, the model used for the induction machine is based on \cite{indm_mod}.  The model is a standard IM model in the synchronous frame with the states selected to be $i_{sd}$, $i_{sq}$, $\psi_{rd}$, $\psi_{rq}$ and $\omega$ which represent the stator current, and the rotor flux linkage in both direct and quadrature axis, and the rotor speed.

\subsection{Parameter Identification Techniques}
For the sake of comparison, the response of the C-PSO identifier is compared with different algorithms namely: Line search technique, GA based technique, global and local PSO techniques. The Line Search (LS) method which is selected to represent the traditional algorithms. GA based technique and  classical PSO  variants fall under the umbrella of EAs.

\subsubsection{GA Based Technique}
The GA used in this application is based on real value representation. The parameters are encoded with real values during initialization to take random values within the bounds given in Table \ref{init_par}.  It is to be noted that the stator and rotor inductances ($L_{sl}$, $L_{rl}$) were combined in a single variable because they are not separable in the induction motor model.  Applications that require explicit knowledge of each of the leakage inductances should perform an extra tests to measure $L_{sl}$.

\begin{table}
  \centering
  \caption{Real values for motor parameters and their corresponding initialization ranges \label{init_par}}
  \begin{tabular}{llll}
    \hline
    Parameter &Real value &Minimum &Maximum\\
    \hline\hline
    $R_s$     &9.203      &1.0     &20.0\\
    $R_r$     &6.61       &1.0     &20.0\\
    $L_{sl}+L_{rl}$ &0.09718 &0.002 &1.0\\
    $L_m$     &1.6816     &0.05    &5.0\\
    $J$       &0.00077    &0.00005 &0.001\\
    \hline
  \end{tabular}
\end{table}

In this application, the SBX recombination operator \cite{deb95} is used due to its strong ability to produce a varied set of offspring which resemble their parents to a certain degree defined by a parameter of this operator ($\eta_c$).  The mutation operator used here is the polynomial mutation \cite{rag01} because it can produce mutations, similar to those produced in binary GA, with a parameter that defines the severity of mutations ($\eta_m$).

The survival selection scheme used here relies on tournament selection to reduce selection pressure and help preserve diversity.  However, instead of copying them to the mating pool, the selected individuals were those who make the population of the next generation.  An elitism strategy is used here to ensure the survival of the most fit individual to prevent a setback in the best found fitness. The parameters of the GA are presented in Table \ref{ga_para}.

\begin{table}
  \centering
  \caption{GA parameters' values for the parameter identification problem \label{ga_para}}
  \begin{tabular}{lll}
    \hline
    Parameter &Description                  &Value\\
    \hline \hline
    $N$       &Population size              &50\\
    $P_c$     &Crossover rate               &0.5\\
    $\eta_c$  &Crossover distribution index &15\\
    $P_m$     &Mutation rate                &0.01\\
    $\eta_m$  &Mutation distribution index  &15\\
    $t_s$     &Tournament size              &2\\
    \hline
  \end{tabular}
\end{table}

\subsubsection{PSO Techniques}
Three variants of the PSO algorithm were used here for parameter identification.  They are based on the local best (PSO-l), global best (PSO-g) topologies, and the C-PSO algorithm.

The particles of the swarm of each one of the three variants are randomly initialized within the initialization ranges of the solution space given in Table \ref{init_par}.  Initial velocities were randomly initialized as well. Just as the case with the other algorithms, the particles were not allowed to fly below the lower bounds of the search space but were allowed to take any value above the upper bound. On the other hand, the velocities were not restricted by any bound.

Based on the corresponding topology used in these variants, the particles are affected by different neighbors and update their positions accordingly.  The parameters of those PSO variants are given in Table \ref{algs_par}.  It is to be noted here that the inertia weight value for the C-PSO algorithm ($w = 1.458$) is twice as much as that for the two other topologies because it is multiplied by a uniformly distributed random number with a mean value of 0.5 leading to an expected value equals to the inertia weights of the two other topologies.

\begin{table}
  \centering
  \caption{PSO parameters' values for the parameter identification problem \label{algs_par}}
  \begin{tabular}{llll}
    \hline
    Algorithm &Parameter &Description            &Value\\
    \hline \hline
    C-PSO     &$N$       &Swarm size             &20\\
              &$\omega$  &Inertia weight         &1.458\\
              &$\chi$    &Constriction coefficient &1\\
              &$\phi_1$  &Personal learning rate &1.494\\
              &$\phi_2$  &Global learning rate   &1.494\\
              &$c_n$     &Number of clubs        &100\\
              &$M_{avg}$ &Average membership     &10\\
              &$M_{min}$ &Min membership level   &4\\
              &$M_{max}$ &Max membership level   &33\\
              &$N$       &Tournament size        &2\\
    \hline
    PSO-l/PSO-g&$N$      &Swarm size             &20\\
              &$\omega$  &Inertia weight         &0.729\\
              &$\chi$    &Constriction coefficient &1\\
              &$\phi_1$  &Personal learning rate &1.494\\
              &$\phi_2$  &Global learning rate   &1.494\\
    \hline
  \end{tabular}
\end{table}

\subsubsection{Line Search}
The LS method is a simple deterministic local search technique.  In this method, the search space is discretized with a unit size of $\delta_i$ for the $i^{th}$ dimension.  A random point ($x$) is chosen in the discretized search space and its fitness is evaluated ($f(x)$), and the fitness of its neighboring points is evaluated as well.  If the fitness value of the most fit neighbor is less than or equal to that of $x$ (for a minimization problem), the algorithm moves to this new point and evaluates the fitness of its neighbors.  But if the fitness of the most fit neighbor is higher than that of $x$, the algorithm terminates.

The set of neighbors $N_x$ for a point $x = (x_1, x_2, \ldots, x_n)$ in a $n$-dimensional space contains $2n$ unique points.  The position of each point in the set can be determined by moving by a step of $\delta_i$ in both directions of the $i^{th}$ dimension; $N_x = (x_1 \pm \delta_1, x_2, \ldots, x_n), (x_1, x_2 \pm \delta_2, \ldots, x_n), \ldots, (x_1, x_2, \ldots, x_n \pm \delta_n)$.

This algorithm contains one parameter which is the step size $\delta_i$.  For the current application, this value was set to $0.1\%$ of the initialization range of the corresponding dimension as explained in Table \ref{init_par}.

\section{Simulation Results}
The five previously mentioned algorithms are used to estimate the real parameters of an induction motor which are given in Table \ref{init_par}.  All the five optimizers used the same fitness function, which evaluates the fitness of the solution passed to it by solving the differential equations based on the parameters of this solution using Matlab's \texttt{ode45} solver and accumulates the error which is the difference between the estimated currents ($\hat{i_1}, \hat{i_2}, \hat{i_3}$) and the measured currents ($i_1, i_2, i_3$). The error value is used as a fitness measure:
\begin{equation}
  f(\theta) = \int_0^T \left(|i_1 - \hat{i_1}| + |i_2 - \hat{i_2}| + |i_3 - \hat{i_3}|\right)dt
\end{equation}
To make a fair comparison between the different optimizers, each one of them was allowed to perform 100,000 function evaluations.  So a larger size for the population or the swarm meant lower number of generations or iterations. The shown results are the average of 10 independent runs of the optimization algorithms.

\begin{figure}
  \centering
  \includegraphics[width=\linewidth]{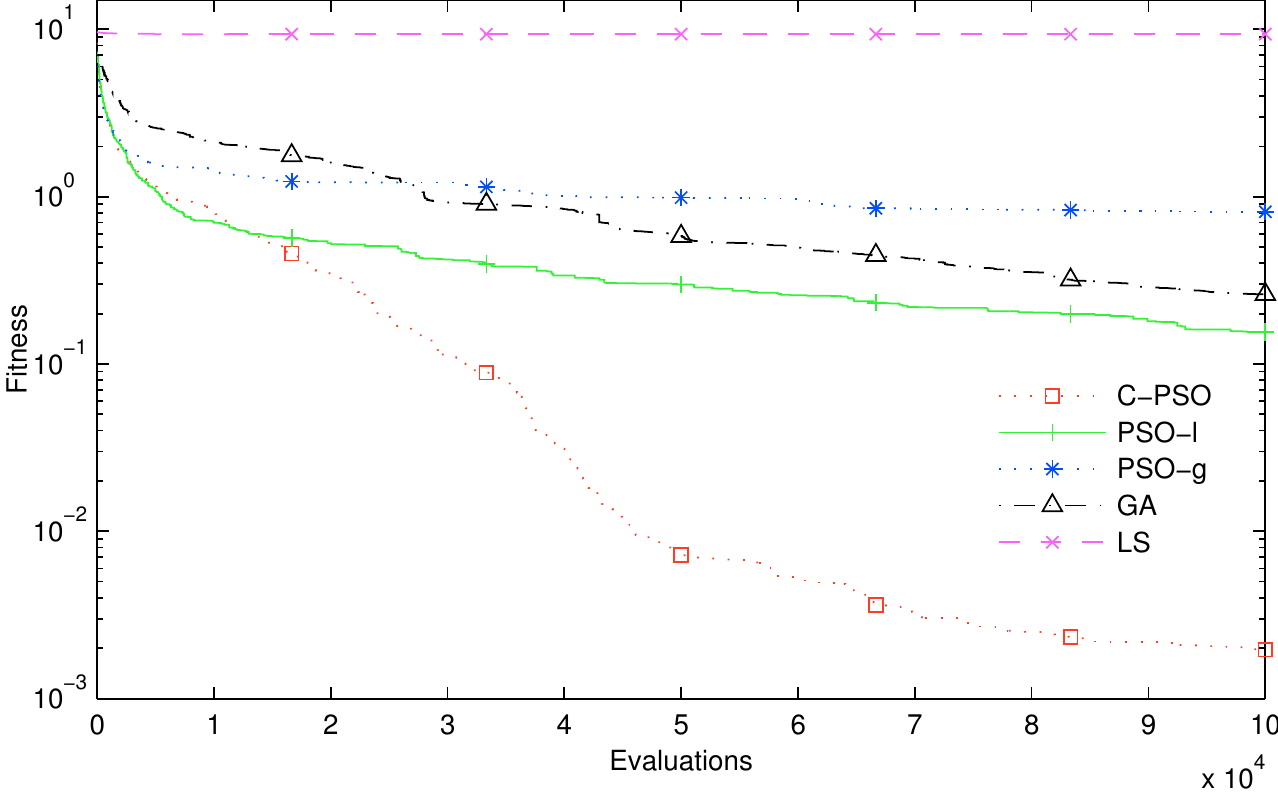}
  \caption{Average fitness obtained by the five optimizers against number of fitness function evaluations \label{conv_graph}}
\end{figure}

Fig. \ref{conv_graph} shows the fitness values obtained by the five optimizers through the 100,000 function evaluations of the simulation.  As can be seen, the C-PSO algorithm reached the lowest fitness value among the five optimizers at the end of the 100,000 function evaluations.  It managed to reach a fitness value of 0.0019 which is significantly better than a value of 0.1554 for the PSO-l algorithm which came in the second place. Little behind the PSO-l algorithm comes the GA then the PSO-g optimizers, while the LS algorithm lags behind them by a long distance.

From the results obtained in Fig. \ref{conv_graph}, the algorithms can be categorized into three groups according to their performance.  The first group contains the LS algorithm which totally lacks any capability of escaping local minima.  In all simulation runs, the algorithm was trapped in the first local minimum it faced, which happened very early in the run (in the first 5,000 function evaluations in most cases), and subsequent evaluations were unnecessary.  In the second group come the PSO-g, GA, and the PSO-l algorithms. The ability of these algorithms to escape local minima is much better than the LS algorithm, which is clear from their lower final fitness values.  Moreover, even at the end of the 100,000 function evaluations, the fitness values of these algorithms were steadily decreasing, but with a small rate.  The third group contains the C-PSO algorithm which outperformed all the other optimizers.

This algorithm shows better ability than the two other groups in escaping local minima.  Moreover, the rate of fitness decrease for this algorithm is much higher than that of all the other algorithms, and it even maintained a reasonable decrease rate at the end of the 100,000 function evaluations. These results show how the C-PSO algorithm exploits the available computation power better than the other algorithms used in our study.

Regarding convergence speed, it is clear from Fig. \ref{conv_graph} that the C-PSO algorithm is the fastest converging algorithm.  By using the number of evaluations needed to reach a value equals 5\% of the initial fitness value (roughly equals 10) as convergence speed measure, it can be seen that the C-PSO algorithm was the fastest converging algorithm among all the optimizers.  After around 15,000 function evaluations it reached the desired fitness value which the PSO-l algorithm achieved after approximately 23,000 function evaluations.  In the third place comes the GA algorithm which needed nearly 60,000 evaluations to reach that fitness value.  However, neither the PSO-g nor the LS algorithms achieved the fitness value in question during the 100,000 function evaluations.  The results obtained here confirms the results obtained in a similar parameter identification study presented in \cite{urs04}, as the PSO algorithms (on average) achieved lower final fitness values and higher convergence speed, which was the case here as well.

Further statistical analysis of the results is shown in Table \ref{fin_fit}.  As can be seen, the C-PSO algorithm achieves the best performance in the four performance measures shown in the Table.  The lowest standard deviation value achieved by the C-PSO algorithm is much lower than that of the PSO-l algorithm, which comes second, shows how the algorithm is more reliable than the other optimizers, because its performance is less dependent on the stochastic variables such as the starting point and the random weight values.  The median value of the different independent runs is a good representative of these runs because it is not affected by the outlier values when compared to their average or mean value.  For this measure, the C-PSO algorithm achieved the lowest fitness value among all the optimizers as well.

\begin{table}
  \centering
  \caption{Final fitness values after 100,000 function evaluations \label{fin_fit}}
  \begin{tabular}{lcccc}
    \hline
    Algorithm &Average        &Std. dev.        &Min.           &Max.\\
    \hline \hline
    C-PSO     &\emph{0.0019}  &\emph{0.0035}    &\emph{2.5e-5}  &\emph{0.0114}\\
    PSO-l     &0.1554         &0.1679           &5.6e-4         &0.5244\\
    PSO-g     &0.8125         &1.3091           &6.5e-4         &4.5732\\
    GA        &0.2607         &0.2250           &2.2e-2         &0.7459\\
    LS        &9.3531         &0.5663           &8.7e-0         &10.4027\\
    \hline

  \end{tabular}
\end{table}

The C-PSO algorithm continues to show superior performance over the other algorithms regarding the average percentage deviation of the estimated parameters from the actual induction motor parameters which are shown in Table \ref{dev_tab}.  It achieves much lower deviation values in three out of the five parameters (by an order of 100 in some situations), and comes second regarding the other two parameters.  As can be seen, the LS algorithm did a bad job in searching for the real parameter values.  The lowest deviation error it achieved, which is approximately 19\%, is an unacceptable error in most real applications (above 5\% deviation error). This deviation error value reached a staggering value of 467\% in the case of the identified inertia value ($J$).

\begin{table}
  \centering
  \caption{Average percentage deviation of the estimated parameters from the real parameters \label{dev_tab}}
  \begin{tabular}{lccccc}
    \hline
            &$R_s$         &$R_r$        &$L_{sl}+L_{rl}$ &$L_m$         &$J$\\
    \hline \hline
    C-PSO   &\emph{0.024}  &1.323        &\emph{0.652}    &\emph{0.029}  &1.684\\
    PSO-l   &1.976         &\emph{1.169} &3.051           &2.188         &2.814\\
    PSO-g   &17.11         &16.21        &25.88           &7.849         &17.69\\
    GA      &3.105         &2.517        &3.349           &0.051         &\emph{0.939}\\
    LS      &19.04         &63.17        &103.4           &28.86         &467.1\\
    \hline
  \end{tabular}
\end{table}

Second worst comes the PSO-g algorithm which achieved deviation errors ranging between 7.8\% and 25.9\%.  Although these values are better than those obtained by the LS algorithm, they are still unacceptable.  Next come the PSO-l and the GA algorithms showing similar performance, however the PSO-l is slightly better as it achieves lower deviation error values in three out of the five parameters being identified.  Those two algorithms achieve a tolerable deviation error (below 5\%) in all parameters.  Ahead of the other optimizers comes the C-PSO algorithm achieving a deviation error lower than 2\% in all five parameters being identified.

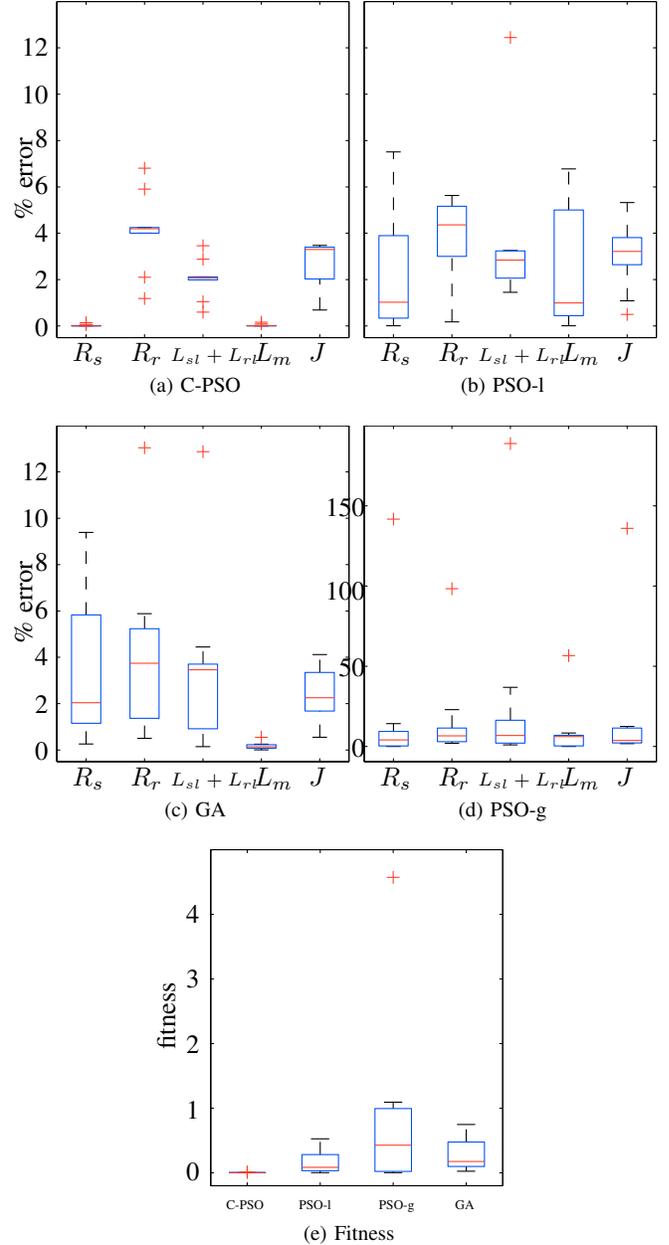
\begin{figure}
  \centering%
  \subfloat[C-PSO]{\label{fig:CSSI_C-PSO_ind_box}
    \begin{overpic}[width=0.45\linewidth]{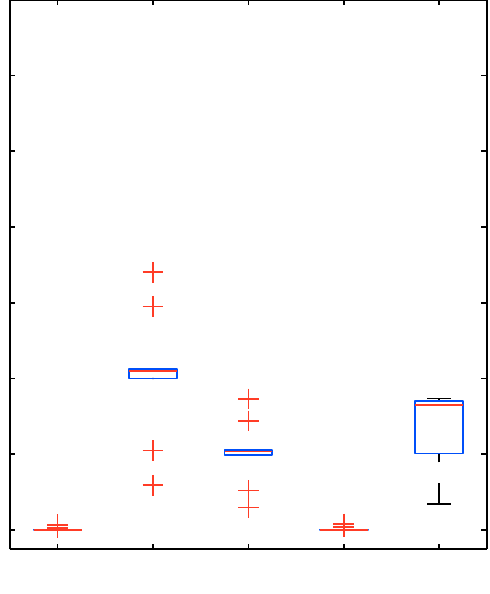}
      \put(6, 1){$R_s$}
      \put(22, 1){$R_r$}
      \put(33, 1){\scriptsize{$L_{sl} + L_{rl}$}}
      \put(56, 1){$L_m$}
      \put(71, 1){$J$}
      \put(-10, 40){\rotatebox{90}{\% error}}
      \put(-8,  8){$\phantom{1}$0}
      \put(-8, 21){$\phantom{1}$2}
      \put(-8, 34){$\phantom{1}$4}
      \put(-8, 47){$\phantom{1}$6}
      \put(-8, 60){$\phantom{1}$8}
      \put(-8, 72){10}
      \put(-8, 85){12}
    \end{overpic}}
  \subfloat[PSO-l]{\label{fig:CSSI_PSOl_ind_box}
    \begin{overpic}[width=0.45\linewidth]{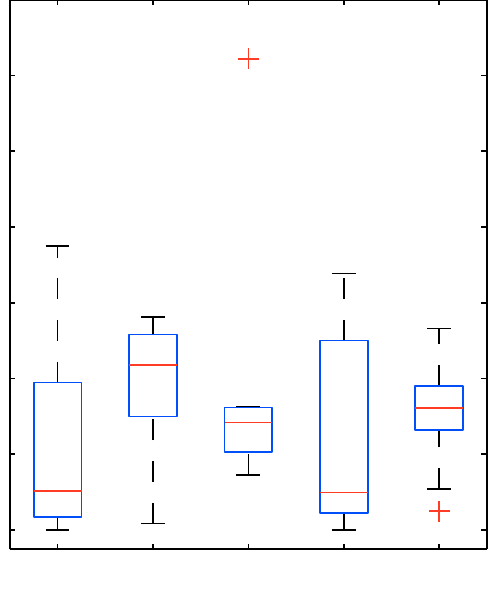}
      \put(6, 1){$R_s$}
      \put(22, 1){$R_r$}
      \put(33, 1){\scriptsize{$L_{sl} + L_{rl}$}}
      \put(56, 1){$L_m$}
      \put(71, 1){$J$}
    \end{overpic}}

  \subfloat[GA]{\label{fig:CSSI_GA_ind_box}
    \begin{overpic}[width=0.45\linewidth]{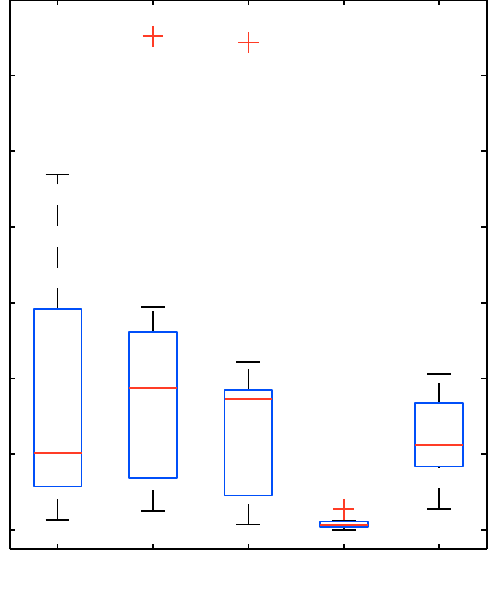}
      \put(6, 1){$R_s$}
      \put(22, 1){$R_r$}
      \put(33, 1){\scriptsize{$L_{sl} + L_{rl}$}}
      \put(56, 1){$L_m$}
      \put(71, 1){$J$}
      \put(-10, 40){\rotatebox{90}{\% error}}
      \put(-8,  8){$\phantom{1}$0}
      \put(-8, 21){$\phantom{1}$2}
      \put(-8, 34){$\phantom{1}$4}
      \put(-8, 47){$\phantom{1}$6}
      \put(-8, 60){$\phantom{1}$8}
      \put(-8, 72){10}
      \put(-8, 85){12}
    \end{overpic}}
  \subfloat[PSO-g]{\label{fig:CSSI_PSOg_ind_box}
    \begin{overpic}[width=0.45\linewidth]{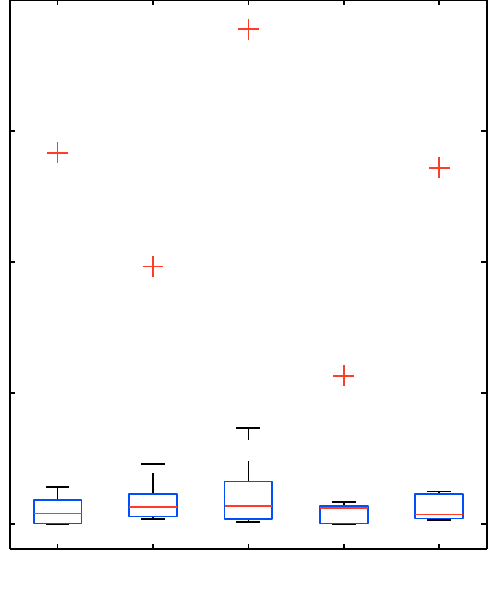}
      \put(6, 1){$R_s$}
      \put(22, 1){$R_r$}
      \put(33, 1){\scriptsize{$L_{sl} + L_{rl}$}}
      \put(56, 1){$L_m$}
      \put(71, 1){$J$}
%      \put(-20, 40){\rotatebox{90}{\% error}}
      \put(-9, 9){$\phantom{15}$0}
      \put(-9, 31){$\phantom{1}$50}
      \put(-9, 52){100}
      \put(-9, 75){150}
    \end{overpic}}

  \subfloat[Fitness]{\label{fig:CSSI_errors_ind_box}
    \begin{overpic}[width=0.45\linewidth]{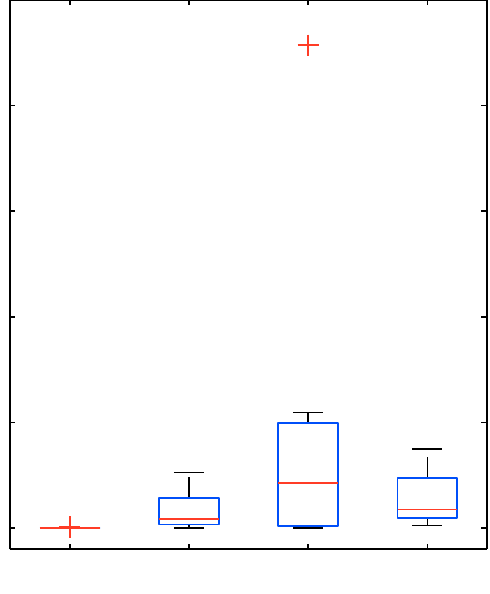}
      \put(6, 1){\tiny{C-PSO}}
      \put(26, 1){\tiny{PSO-l}}
      \put(48, 1){\tiny{PSO-g}}
      \put(69, 1){\tiny{GA}}
      \put(-13, 45){\rotatebox{90}{fitness}}
      \put(-5, 9){0}
      \put(-5, 26){1}
      \put(-5, 44){2}
      \put(-5, 62){3}
      \put(-5, 80){4}
    \end{overpic}}
  \caption{Boxplots showing the performance of the different AI algorithms \label{boxplot}}
\end{figure}

Further statistical analysis of the obtained results is presented in Fig. \ref{boxplot} using boxplots.  First, Fig. \ref{boxplot}(a)--(d) show statistical data regarding the estimated parameters.  As can be seen, Fig. \ref{boxplot}(a) shows how the C-PSO had more restricted outliers (lower deviation from the mean) when compared to the other algorithms shown in Fig. \ref{boxplot}(b)--(d).  Moreover, the deviation from the mean is graphically shown to be less in the case of the C-PSO algorithm than in the case of the other optimizers.  Fig. \ref{boxplot}(e) shows statistical data regarding the final fitness values obtained by the four optimizers.  Again, C-PSO shows superior performance;  The obtained results are very close to the mean value and there are no outliers compared to the other optimizers with higher deviation from the mean and more outliers.

\begin{figure}
  \centering
  \includegraphics[width=\linewidth]{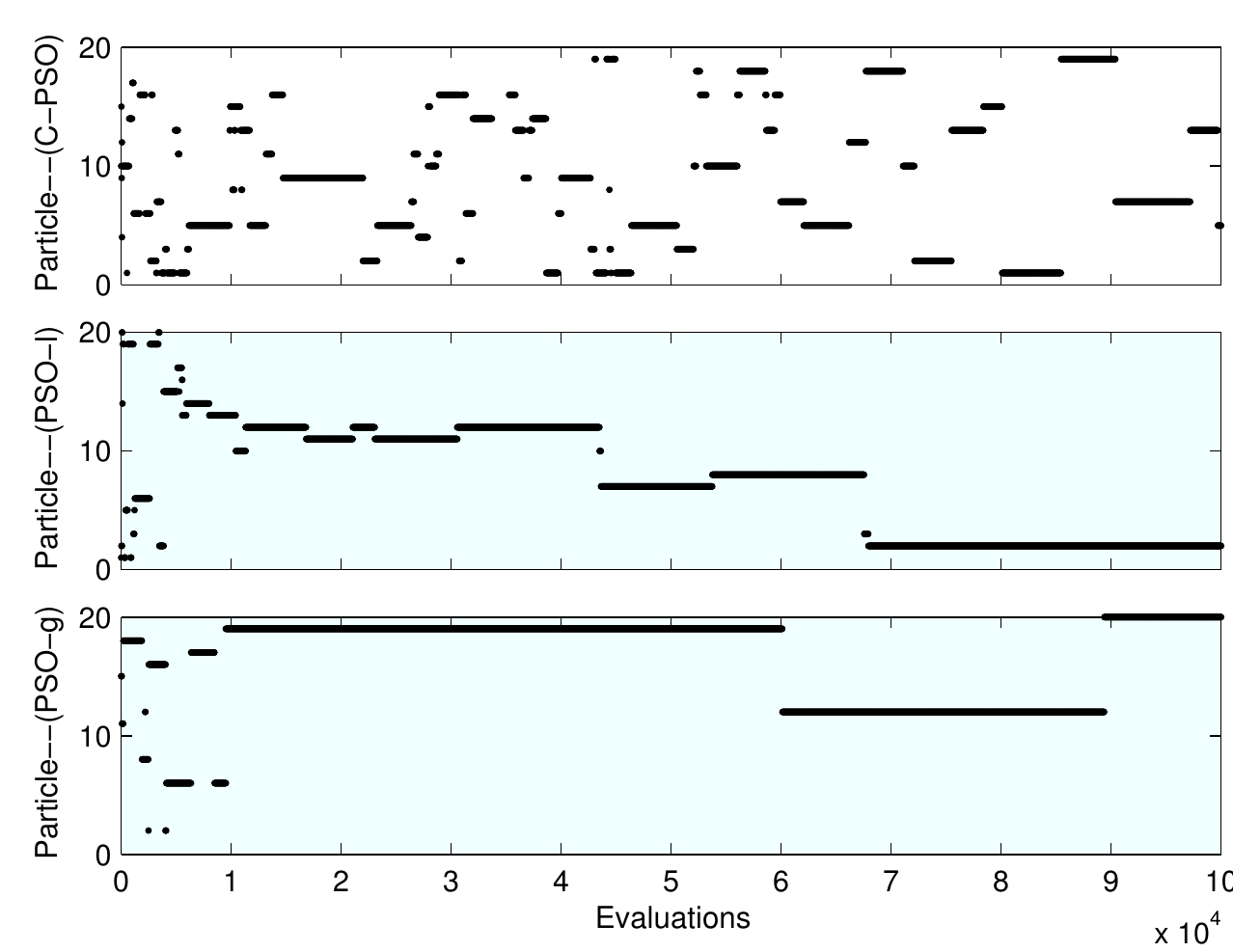}
  \caption{Best particle in the swarm of the three PSO algorithms used in parameter identification, C-PSO (top), PSO-l (middle), and PSO-g (bottom) \label{particles}}
\end{figure}

Fig. \ref{particles} presents the index values of the best performing particle in the swarm for the three PSO algorithms against number of fitness function evaluations. As can be seen, most of the particles in the C-PSO swarm participated in the search process as the status of the best particle in the swarm was alternating among almost all the particles (Fig. \ref{particles} up).  On the other hand, the status of the best particle in the PSO-l algorithm was confined to fewer particles (Fig. \ref{particles} middle), and each one of them claimed that status for a longer period of time (on average) than the case of the C-PSO algorithm.  The effect of the ring topology is clear in this case as best particle status moves from a particle to its neighbor in the ring. Finally the behavior of the particles of the PSO-g algorithm is shown in Fig. \ref{particles}--bottom.  Only three particles (\#13, \#19, and \#20) were leading the swarm in the last 90,000 function evaluations.

The alternation of the best particle status as depicted in the C-PSO case shows that most of the particles of the swarm participated effectively in the search process; While some particles are searching for the global optimum in one region, the other particles are searching for that optimum elsewhere, but are guided by the experience of the other particles in the swarm. This effective search mechanism was present but with less efficiency in the case of the PSO-l algorithm, and this efficiency is much more less in the case of the PSO-g as few particles are effectively searching for the global minimum while the others are being dragged by them.

\section{Conclusions}
Particle swarm optimizers are very sensitive to the shape of their social network.  Both PSO-g and PSO-l lack the ability of adapting their social network to the landscape of the problem they work on.

The proposed C-PSO algorithm overcomes this problem. The dynamic social network of the optimizer shrinks the social network of superior particles to reduce their influence on other particles, while expanding it for the worst particles to increase their chance in learning from better particles.

C-PSO achieved better results than PSO-l and PSO-g either in escaping local optima or in convergence speed to global optimum.

Further investigations have shown that the dynamic social network allowed particles to be guided indirectly by the superior particles, while searching for better solutions more freely than the case of PSO-g.  It was shown using empirical\newpage\noindent results that C-PSO is able to explore and find better regions in the search space during periods of stagnation, making it attractive for use in multimodal problems.

\bibliography{biblio}
\bibliographystyle{IEEEtran}

\end{document}